\documentclass[10pt,twocolumn,letterpaper]{article}
\usepackage{placeins}
\usepackage{float}

\usepackage{multirow}

\PassOptionsToPackage{hyphens}{url} % just add for arxiv

\PassOptionsToPackage{table}{xcolor}
\usepackage[pagenumbers]{cvpr} % To force page numbers, e.g. for an arXiv version
\definecolor{cvprblue}{rgb}{0.21,0.49,0.74}
\usepackage[pagebackref,breaklinks,colorlinks,allcolors=cvprblue]{hyperref}

\def\confName{CVPR}
\def\confYear{2027}

\title{TRUST-Stereo: Teacher-Assisted Robust Uncertainty-Guided Search and Transfer for Fast Stereo Matching}

\author{
Pouya Sohrabipour \qquad
Yihong Feng \qquad
Chaitanya Kumar Reddy Pallerla\\
Tuan Dang \qquad
Ngan Le \qquad
Dongyi Wang\\
University of Arkansas
}

\begin{document}
\maketitle

\begin{abstract}
Coarse-to-fine stereo matching enables efficient disparity
estimation but remains sensitive to errors in propagated
disparity, which may place ground-truth correspondences
outside subsequent local search ranges. We propose
TRUST-Stereo \textbf{T}eacher-Assisted \textbf{R}obust
\textbf{U}ncertainty-Guided \textbf{S}earch and
\textbf{T}ransfer for fast  \textbf{Stereo}, 
that addresses this limitation through an
Uncertainty-Guided Residual Search Module (UGRSM).
UGRSM estimates the reliability of propagated disparity
and predicts residual offsets to adaptively relocate
local cost-volume centers. Additionally, we introduce
a two-stage Senior--Junior Teacher Assistant knowledge
distillation strategy to transfer rich monocular
representations from a large foundation model to a
lightweight backbone to achieve fast stereo matching performance. Experiments on KITTI 2012,
KITTI 2015, ETH3D, and Middlebury demonstrate
strong zero-shot generalization. Using only 1.4M
synthetic stereo pairs, TRUST-Stereo achieves
state-of-the-art results on several benchmark metrics,
outperforming competing efficient methods trained
with substantially larger datasets, including
real-world pseudo-labeled samples.
Specifically, it achieves a KITTI 2015 EPE of 0.93, ETH3D Bad 1.0 of 2.13\% and Middleblury Bad 2.0 of 6.79\% while maintaining
 inference at approximately 50 ms per
stereo pair.
\footnote{\url{https://github.com/PouyaSohrabipour/TRUST-Stereo}}
\end{abstract}

%%%%%%%%% BODY TEXT
\section{Introduction}
\label{sec:intro}
\begin{figure}[t]
    \centering
    \includegraphics[width=\columnwidth]{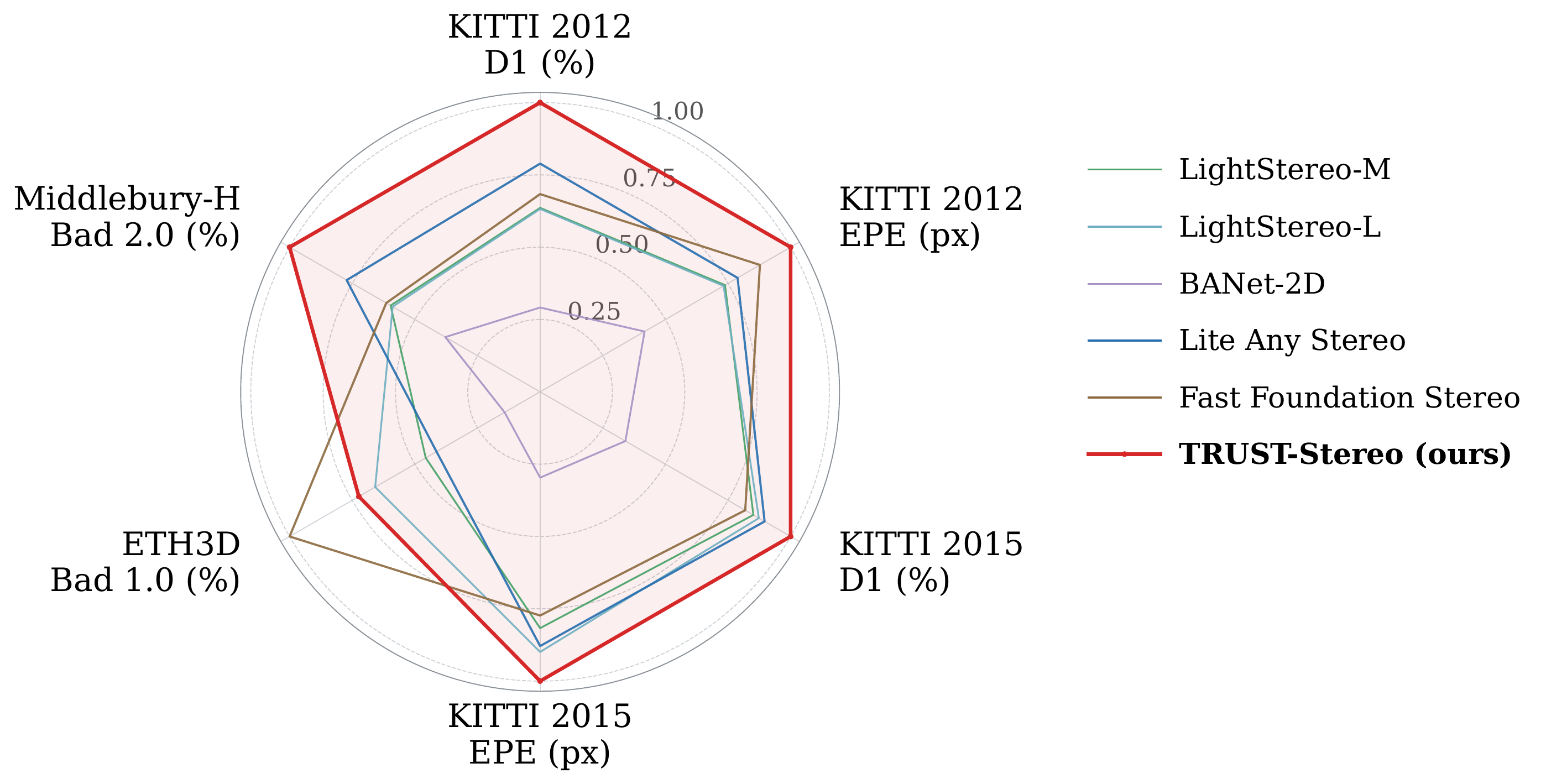}
    \caption{Zero-shot generalization comparison under
    SceneFlow-only training.}
    \label{fig:radar}
\end{figure}
Stereo matching is widely used for dense depth estimation due
to its strong geometric foundation. Recent learning-based
methods achieve high accuracy through cost-volume construction
and disparity regression \cite{scharstein2002taxonomy,tosi2025survey}. However, dense cost volumes at high
resolutions and large disparity ranges incur substantial
computational and memory costs, limiting the real-time
deployment~\cite{xu2023iterative,wen2025foundationstereo,liu2024goat,wei2025waveletstereo,wang2025promptstereo,chen2024mochastereo}.

To improve reconstruction accuracy and efficiency, recurrent stereo matching architectures,
such as RAFT-Stereo~\cite{lipson2021raft} and
IGEV-Stereo~\cite{xu2023iterative}, have demonstrated
that Gated Recurrent Unit (GRU)-based iterative
refinement can progressively correct initial disparity
errors and improve matching accuracy. Coarse-to-fine stereo methods are cluster of models that can estimate
disparity at low resolution and progressively refine the disparity estimation using
local cost volumes to improve the match efficiency ~\cite{cheng2025monsterplusplus,lipson2021raft,wang2019anytime}.

However, integrating the two ideas are still challenging to maintain both computational accuracy and efficiency. Coarse-to-fine stereo methods, although efficient, rely heavily on accurate disparity
propagation. Errors introduced during coarse estimation or
upsampling may place the ground-truth correspondence outside
the local search range, preventing subsequent refinement
stages from recovering the correct disparity. Such failures
are particularly common near object boundaries, thin
structures, occlusions, and large disparity discontinuities \cite{tankovich2021hitnet,liu2024goat,mao2021uasnet,shen2021cfnet}. 

% To mitigate these failures, previous work explored different uncertainty-aware 
% adaptive disparity search methods. UASNet~\cite{mao2021uasnet} estimates uncertainty from cost-volume probability distributions. CREStereo++~\cite{jing2023uncertainty} employs
% variance-based uncertainty derived from correlation
% responses. Both method then use the estimated uncertainty to refine the search in ambigous region for more confident predictions. However, these designs cannot naturally fit for the coarse-to-fine recurrent architectures, where disparity
% estimates undergo multiple refinement iterations
% and subsequent upsampling before being propagated
% to finer scales. Consequently, uncertainty estimated from earlier
% matching distributions may no longer accurately
% represent the reliability of the refined and propagated
% disparity. More importantly, such uncertainty
% primarily characterizes matching ambiguity rather
% than explicitly determining whether propagation
% errors place the ground-truth correspondence
% outside the next local search window.

To mitigate these failures, previous works have explored uncertainty-aware adaptive stereo matching. UASNet~\cite{mao2021uasnet} leverages cost-volume probability distributions to predict adaptive disparity ranges and sampling locations, while CREStereo++~\cite{jing2023uncertainty} employs variance-based uncertainty to guide content-aware feature warping during recurrent refinement. However, uncertainty derived from matching distributions primarily characterizes matching ambiguity rather than the actual reliability of disparity estimates. Specifically, a strong but incorrect matching response may produce low uncertainty even when the ground-truth correspondence lies outside the local search window. Conversely, ambiguous or multimodal matching responses may produce high uncertainty despite the ground-truth correspondence already being within the search range. Such discrepancies can lead to insufficient correction of inaccurate search centers or unnecessary adaptation of reliable ones. This limitation is particularly important in coarse-to-fine recurrent architectures, where disparity estimates undergo iterative refinement and are subsequently propagated across scales, potentially shifting the next local search window away from the correct correspondence.

To address this issue, we propose UGRSM, which adaptively relocates local cost-volume search centers while maintaining a fixed search radius. The module learns uncertainty through pseudo-ground-truth supervision derived from the propagated disparity error relative to the next-stage search radius. Unlike uncertainty derived solely from matching statistics, our formulation explicitly supervises the uncertainty predictor to identify whether the propagated disparity error exceeds the local search range. This enables the model to distinguish unreliable search centers requiring correction from those already covering the correct correspondence. The predicted uncertainty then selectively guides residual offsets to relocate local cost-volume centers, improving correspondence coverage without increasing the search radius or number of disparity candidates.

Addressing these limitations requires improving two
complementary aspects of coarse-to-fine stereo matching:
the quality of extracted image features and the reliability
of subsequent disparity refinement. Recent monocular foundation models provide rich
semantic and structural representations that can
enhance stereo feature extraction, particularly in
textureless, repetitive, and occluded regions
\cite{depth_anything_v2,oquab2023dinov2,
cheng2025monster,bartolomei2025stereoanywhere}.
However, large backbones such as Depth Anything V2
ViT-L introduce substantial inference costs
\cite{depth_anything_v2,jing2026lite}.
Lightweight alternatives improve efficiency but
introduce a representation capacity gap that
complicates knowledge transfer
\cite{mirzadeh2020improved}. In addition, even with strong features,
errors remaining after recurrent refinement and
upsampling may shift the next local search window
away from the correct correspondence.

To address these challenges, we propose
\textbf{T}eacher-Assisted \textbf{R}obust
\textbf{U}ncertainty-Guided \textbf{S}earch and
\textbf{T}ransfer for fast  \textbf{Stereo}
Matching (TRUST-Stereo), a real-time coarse-to-fine
stereo matching framework that combines
teacher-assisted feature extraction with
uncertainty-guided disparity search.

First, we introduce a two-stage Senior--Junior
Teacher Assistant (SJTA) distillation strategy
to enhance the feature extraction capability and efficiency
of a lightweight ViT-S encoder. Existing Teacher Assistant Knowledge Distillation
(TAKD)~\cite{mirzadeh2020improved} offers an good way to distillate teacher knowledge through an intermediate assistant. However, it requires a large-scale retraining on millions of images for any new datasets applied which can incur substantial computational overhead. In the first stage of our design, pretrained DINOv2-based
ViT-L (Teacher) and ViT-S (Junior TA) backbones are employed to
construct a teacher-enriched intermediate
representation through cross-attention and
self-attention. The intermediate representation is upgraded version of Junior TA, and therefore we call it Senior TA. 
This stage is performed only once and is
never repeated during subsequent dataset-specific fine-tuning.
In the second stage, the Senior TA can independently
supervises the lightweight student through
feature-level distillation.
Thus, although SJTA employs two training stages,
only the second stage is repeated for subsequent
training or adaptation, avoiding the computational
overhead of retraining the intermediate assistant in traditional TAKD
while improving feature representations for
stereo correspondence estimation.

Second, to guarantee reliable disparity
propagation in the course-to-fine architecture, we introduce
an Uncertainty-Guided Residual Search Module
(UGRSM). Instead of using probability estimation, the module directly estimates propagated disparity
reliability relative to the next local search
range and predicts uncertainty-guided residual
offsets to adaptively relocate local cost-volume
centers before refinement.

Together, SJTA and UGRSM address complementary
sources of error in coarse-to-fine stereo matching:
SJTA strengthens efficient feature representations for
correspondence estimation, while UGRSM improves
the robustness of local search against remaining
disparity propagation errors. Their integration
supports accurate zero-shot stereo matching
while preserving real-time inference.

As illustrated in Fig.~\ref{fig:radar},
TRUST-Stereo demonstrates strong zero-shot
generalization under SceneFlow-only training,
achieving the best performance on five of six
evaluation metrics among the compared SOTA efficient
stereo matching methods across KITTI 2012~\cite{geiger2012kitti},
KITTI 2015~\cite{menze2015kitti},
ETH3D~\cite{schops2017eth3d},
and Middlebury-H~\cite{scharstein2014middlebury}.

The main contributions are summarized as follows:

\begin{itemize}

\item We identify a key limitation of coarse-to-fine stereo
matching, where disparity propagation errors shift the
local search range away from the ground-truth correspondence.

\item We propose UGRSM, which estimates propagated disparity
reliability and predicts residual offsets to adaptively
relocate local cost-volume centers.

\item We introduce a two-stage Senior--Junior Teacher Assistant
distillation strategy that transfers foundation-model
representations to a lightweight backbone without separate
large-scale teacher-assistant pretraining. 

\item We demonstrate state-of-the-art (SOTA) zero-shot
generalization, outperforming competing efficient stereo
methods on five of six metrics under SceneFlow-only
pretraining~\cite{mayer2016dispnet}. With only 1.4M
synthetic stereo pairs, TRUST-Stereo achieves SOTA
performance across all six evaluated metrics, including
a KITTI 2015 D1 of 3.49\% and ETH3D Bad 1.0 of
2.13\%, while maintaining fast inference at approximately
50 ms per stereo pair. Furthermore, benchmark-specific
fine-tuning achieves competitive results on KITTI
2012~\cite{geiger2012kitti} and a D1-all of 1.64\%
on KITTI 2015~\cite{menze2015kitti}.
\end{itemize}

%-------------------------------------------------------------------------
\begin{figure*}[t]
\centering
\includegraphics[width=\textwidth]{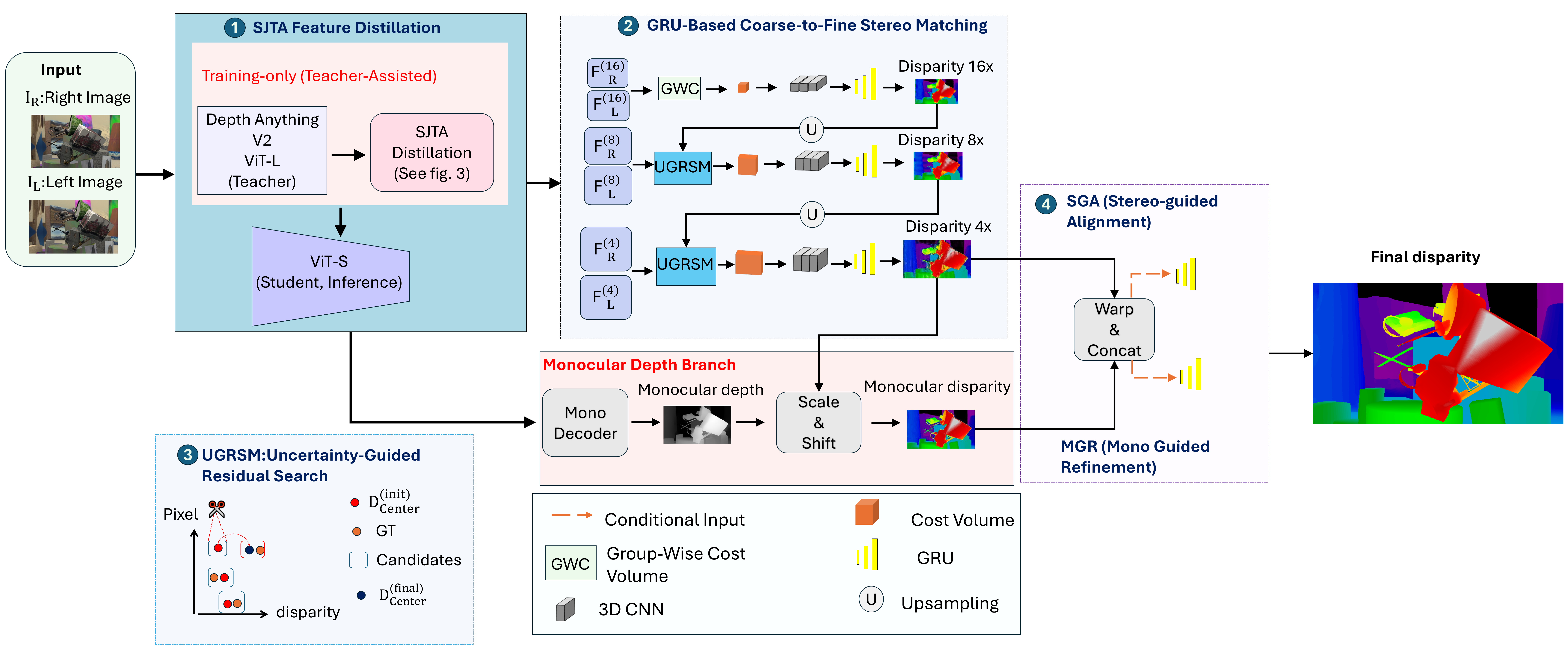}

\caption{Overview of TRUST-Stereo. A coarse-to-fine
stereo matching architecture iteratively refines disparity
using ConvGRUs. The proposed UGRSM adaptively adjusts
local search centers, while Senior--Junior Teacher Assistant
distillation provides lightweight monocular guidance
for robust zero-shot generalization.
%\cm{Check data/control flow. Example: (3) has only in-flow}
} 

\label{fig:main}
\end{figure*}

\section{RELATED WORKS }

Deep depth estimation methods fall into three groups: optical triangulation/stereo matching methods, monocular depth estimation methods, and hybrid mono--stereo approaches. Each category addresses depth estimation from a different perspective, with distinct trade-offs between accuracy, efficiency, and robustness.

\subsection{Stereo Matching Based Depth Estimation}

Stereo matching estimates depth by exploiting geometric correspondences between rectified image pairs. Early-stage deep learning approaches replaced handcrafted matching costs with learned representations, demonstrating the effectiveness of convolutional neural networks for disparity estimation 
\cite{kendall2017end,zhang2019ganet,shen2021cfnet,mao2021uasnet,zhao2023high,feng2023mc,xu2023unifying,zbontar2015mccnn,mayer2016dispnet}.
These methods construct a cost volume over disparity candidates and regress dense disparity maps end to end. 

Subsequent works focused on improving accuracy through cost-volume
regularization using 3D convolutional neural networks
\cite{guo2019gwcnet,bangunharcana2021correlate,cheng2020hierarchical,
duggal2019deeppruner,wen2026fastfoundationstereo}.
Pyramid Stereo Matching Network (PSMNet) \cite{Chang_Che}
introduced spatial pyramid pooling to enlarge the receptive field,
while later methods such as GA-Net \cite{zhang2019ganet}
and cascade-based architectures
\cite{gu2020cascade}
further enhanced geometric consistency. Additionally, recurrent approaches such as
RAFT-Stereo~\cite{lipson2021raft} and
IGEV-Stereo~\cite{xu2023iterative} employ
GRU-based iterative updates to progressively
refine disparity estimates using correlation
and cost-volume features. Despite
these advances, stereo matching remains challenging in
textureless, repetitive, and occluded regions, where
ambiguous correspondences can lead to inaccurate
disparity estimation.

Recent efforts have specifically focused on real-time stereo
matching by adopting coarse-to-fine architectures that
restrict disparity search to local cost volumes centered
on upsampled estimates~\cite{wang2019anytime,gu2020cascade,
cheng2025monsterplusplus}.
AnyNet~\cite{wang2019anytime} progressively refines
disparity through residual matching at finer resolutions,
while RT-MonSter++~\cite{cheng2025monsterplusplus}
employs global matching at the coarsest scale and
local cost volumes at finer scales to reduce computation.
However, these approaches rely on accurate disparity
propagation between scales. When propagation errors
exceed the local search radius, the ground-truth
correspondence falls outside the search window,
limiting the ability of subsequent refinement stages
to recover the correct disparity.

\subsection{Monocular Depth Estimation}
Unlike stereo matching, which relies on reliable
correspondences between image pairs, monocular depth
estimation leverages learned semantic and contextual
priors to infer depth, offering advantages in textureless
and repetitive regions where stereo matching becomes
ambiguous.
Monocular depth estimation predicts scene depth from a single RGB image by learning semantic and contextual cues such as perspective, object size, and texture gradients \cite{godard2017unsupervised,monodepth2, Ranftl2022, Birkl2023MiDaSv31, depth_anything_v2}. 

% Early deep learning models employed multi-scale convolutional architectures to regress dense depth maps \cite{eigen2014depth,laina2016deeper}. Unsupervised and self-supervised approaches later leveraged stereo consistency during training to reduce reliance on ground-truth depth annotations \cite{godard2017unsupervised}.

Recently, large-scale pretraining and vision foundation models
have significantly advanced monocular depth estimation.
Depth Anything V2~\cite{depth_anything_v2} offers both large
(ViT-L) and small (ViT-S) backbones. While ViT-L provides
rich semantic and structural representations that can enhance
stereo matching, its high computational cost limits real-time
deployment. In contrast, ViT-S offers greater efficiency
but has lower representation capacity, potentially reducing
feature quality in challenging regions. This introduces a
trade-off between feature representation quality and
computational efficiency. Despite their strong
semantic understanding, monocular methods often struggle
with metric scale ambiguity for reliable robotics control
applications~\cite{Bhat2023ZoeDepth,tri-zerodepth,Zeng2024RSA}.

\subsection{Hybrid Mono--Stereo Methods}

Hybrid methods combine the geometric accuracy of stereo
matching with the semantic robustness of monocular depth
estimation. MonSter~\cite{cheng2025monster} integrates
monocular and stereo disparity predictions through dual
refinement branches, using flaw maps and cross-modal
guidance to correct unreliable regions. However, its
multi-stage architecture introduces substantial
computational overhead.

\section{Our Approach}

\subsection{Overview}

We propose TRUST-Stereo, a fast coarse-to-fine stereo
matching framework illustrated in Fig.~\ref{fig:main}. The proposed two-stage SJTA
knowledge distillation strategy transfers representations
from a large pretrained teacher to a lightweight backbone.
In the first stage, cross-attention and self-attention construct
a teacher-enriched Junior TA representation. In the second
stage, the intermediate Junior TA becomes the Senior TA
and supervises the lightweight student, enabling efficient inference without the large teacher. Additionally, the proposed Uncertainty-Guided Residual Search
Module (UGRSM) estimates the reliability of propagated disparities
and predicts residual offsets to adaptively relocate local
cost-volume centers, improving disparity refinement.

\subsection{Senior--Junior Teacher Assistant Feature Distillation}

\begin{figure*}[t]
\centering
\includegraphics[width=\textwidth]{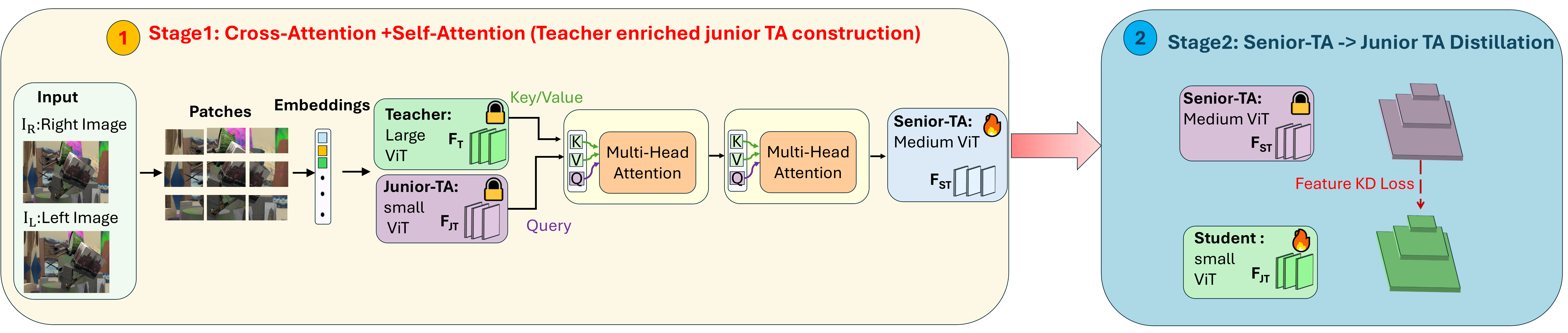}
\caption{Overview of the proposed two-stage Senior--Junior Teacher Assistant knowledge distillation framework.}
\label{fig:KD}
\end{figure*}

TRUST-Stereo introduces a two-stage SJTA distillation framework to transfer rich representations from a pretrained Depth Anything V2 ViT-L teacher to a lightweight ViT-S student, as illustrated in Fig.~\ref{fig:KD}.

Direct distillation is hindered by the substantial teacher--student capacity gap. Conventional teacher assistant methods can introduce considerable training overhead, but it still relies on a large teacher model for adapting new datasets or tasks. To address the limitations, SJTA leverages pretrained ViT-L (Teacher) and ViT-S (Junior TA) backbones to construct an intermediate teacher assistant (Senior TA) using only one dataset (SceneFlow). This first stage is performed only once, while subsequent training or adaptation, including million-scale training,  requires only the second stage, avoiding redundant large teacher model involvement in following task.Here, million-scale training refers to training on approximately 1.4 million synthetic stereo pairs from multiple datasets. To train the Senior TA, cross-attention integrates Junior TA features as queries and teacher features as keys and values, followed by self-attention to learn a teacher-enriched Senior TA representation.

In the second stage, the trained Senior TA supervises the lightweight ViT-S student through feature-level knowledge distillation:

\begin{equation}
\mathcal{L}_{KD}^{(l)}
=
\left\|
F_{\mathrm{ST}}^{(l)}
-
F_{\mathrm{Student}}^{(l)}
\right\|_2^2,
\end{equation}

where $F_{\mathrm{ST}}^{(l)}$ and $F_{\mathrm{Student}}^{(l)}$ denote the Senior TA and student features at level $l$, respectively. Distillation is applied across selected feature levels. During inference, only the lightweight ViT-S student is retained, introducing no additional distillation-related computational overhead.

\subsection{GRU-Based Coarse-to-Fine Stereo Matching}

TRUST-Stereo employs three refinement stages at
$\frac{1}{16}$, $\frac{1}{8}$, and $\frac{1}{4}$
of the input resolution
~\cite{cheng2025monsterplusplus,lipson2021raft,xu2023iterative}.
At the coarsest $\frac{1}{16}$ scale,
a full group-wise correlation (GWC) cost volume
is constructed from left and right image features:

\begin{equation}
C_g(d,x,y)
=
\frac{1}{N_g}
\left\langle
F_{L,g}^{(16)}(x,y),
F_{R,g}^{(16)}(x-d,y)
\right\rangle,
\end{equation}

where $(x,y)$ denotes spatial coordinates,
$d$ is a candidate disparity at the
$\frac{1}{16}$ scale, and $g$ indexes
the $G$ channel groups.
$F_{L,g}^{(16)}$ and $F_{R,g}^{(16)}$
denote the left and right feature vectors
of group $g$, respectively,
$N_g$ is the number of channels per group,
and $\langle\cdot,\cdot\rangle$
denotes the inner product.
GWC measures the similarity between left-
and right-image features at each candidate
disparity, with higher correlation indicating
stronger evidence of a stereo correspondence.

The cost volume is regularized using 3D
convolutions, followed by softmax normalization
and soft-argmin disparity regression to obtain
the initial disparity estimate.

% Following~\cite{lipson2021raft,xu2023iterative},
% Convolutional Gated Recurrent Units (ConvGRUs)
% iteratively refine disparity at each scale.
% At iteration $i$ and scale $s$, where
% $s\in\{16,8,4\}$ denotes the downsampling factor.
% Correlation features are sampled using the
% current disparity estimate. Convolutional layers
% processes the sampled correlation features and
% the current disparity estimate together
% with its recurrent hidden state to predict
% a residual disparity correction $\Delta D_i^{(s)}$, and$D_{i+1}^{(s)} = D_i^{(s)}+\Delta D_i^{(s)}$

Following~\cite{lipson2021raft,xu2023iterative},
Convolutional Gated Recurrent Units (ConvGRUs)
iteratively refine disparity at each scale.
At iteration $i$ and scale $s$, where
$s\in\{16,8,4\}$ denotes the downsampling factor,
the current disparity estimate $D_i^{(s)}$ is used
to determine the sampling locations in the correlation
volume. The sampled correlation features are then passed to
the ConvGRU along with the current disparity estimate and
recurrent hidden state to predict a residual disparity
correction $\Delta D_i^{(s)}$. The disparity is updated as

\begin{equation}
D_{i+1}^{(s)} = D_i^{(s)} + \Delta D_i^{(s)}.
\end{equation}

Here, $D_i^{(s)}$ and $D_{i+1}^{(s)}$ denote the disparity
estimates before and after iteration $i$ at scale $s$,
respectively.

The refined disparity and interpolated hidden
state are propagated to the next finer scale.
At $\frac{1}{8}$ and $\frac{1}{4}$ resolution,
the proposed UGRSM adjusts the local search
centers before constructing the corresponding
local cost volumes, as detailed in the
following section.

\subsection{Uncertainty-Guided Residual Search Module}
In coarse-to-fine stereo matching, local cost volumes
search within a limited disparity range centered on
the propagated disparity estimate. When propagation
errors exceed this range, the correct correspondence
falls outside the search window, limiting the ability
of subsequent refinement to recover it.
Expanding the search range increases computational
cost. To identify unreliable search centers,
we define uncertainty based on the propagated
disparity error relative to the local search radius. Based on this definition, UGRSM learns to predict uncertainty
from the propagated disparity and image features, and the
predicted uncertainty can guide residual offsets that adaptively
adjust local cost-volume centers.

\begin{figure}[t]
\centering
\includegraphics[width=\linewidth]{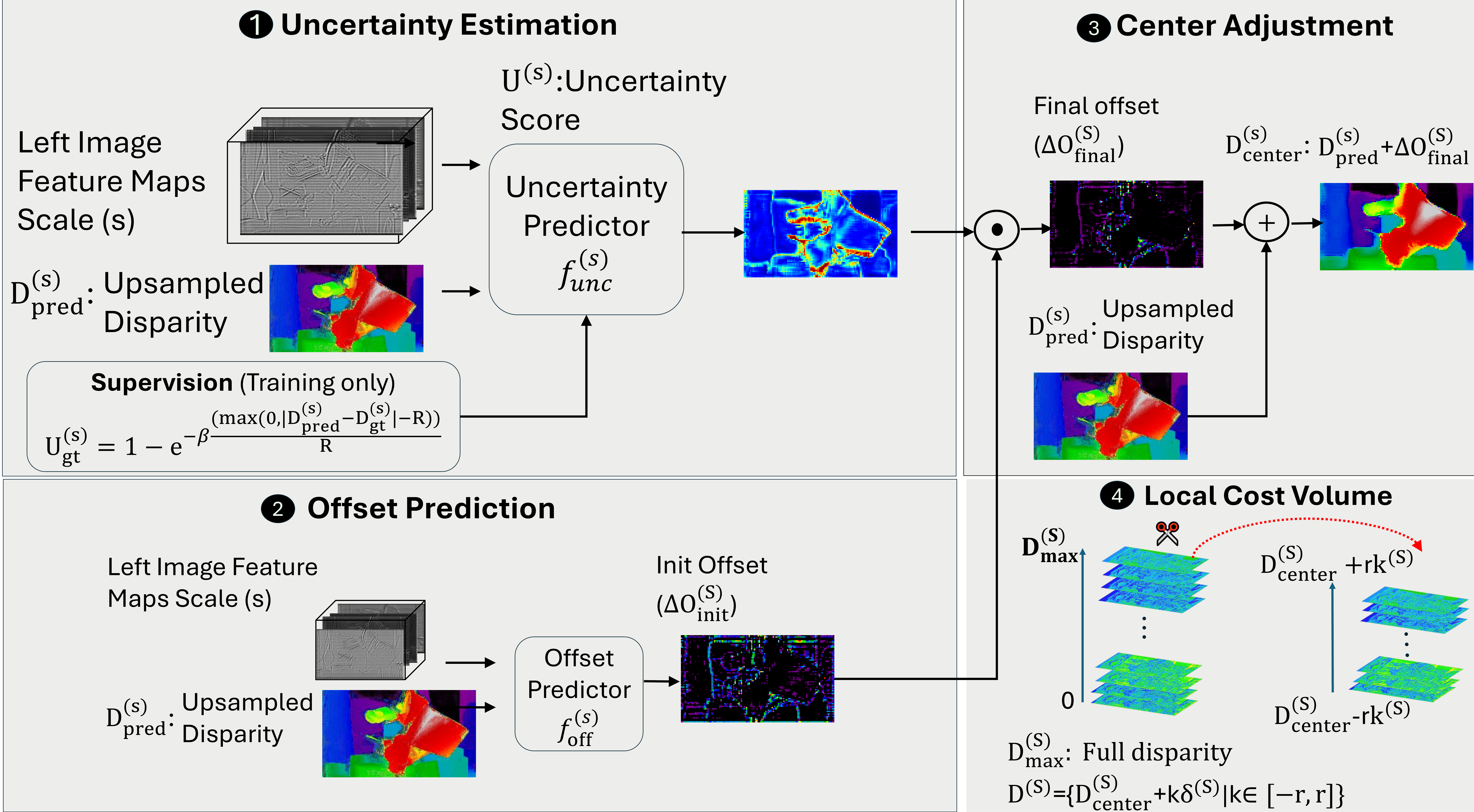}
\caption{
Architecture of the proposed Uncertainty-Guided Residual Search Module (UGRSM). The module predicts an uncertainty map and a residual search offset from the propagated disparity and image features. The uncertainty score modulates the residual offset to produce an adjusted local cost-volume center used for local disparity search.
%\cm{Same comments: texts are too small}
}
\label{fig:ugrsm}
\end{figure}

% To start, UGRSM predicts an uncertainty map $U^{(s)}\in[0,1]$
% and an initial residual offset $\Delta O_{init}^{(s)}$
% from the left-image features $F_L^{(s)}$ and the
% propagated disparity $D_{pred}^{(s)}$. $U^{(s)} =
% f_{unc}^{(s)}(F_L^{(s)},D_{pred}^{(s)})$ and 
% $\Delta O_{init}^{(s)} =
% f_{off}^{(s)}(F_L^{(s)},D_{pred}^{(s)})$. The offset predictor $f_{off}^{(s)}$ employs two
% $3\times3$ convolutions followed by two $1\times1$
% convolutions. For training $U^{(s)}$, since ground-truth uncertainty labels
% are unavailable, pseudo-targets are generated from
% the propagated disparity error:

To start, UGRSM predicts an uncertainty map $U^{(s)}\in[0,1]$
and an initial residual offset $\Delta O_{init}^{(s)}$
from the left-image features $F_L^{(s)}$ and the
propagated disparity $D_{pred}^{(s)}$:

\begin{equation}
\begin{aligned}
U^{(s)} &=
f_{unc}^{(s)}(F_L^{(s)},D_{pred}^{(s)}),\\[4pt]
\Delta O_{init}^{(s)} &=
f_{off}^{(s)}(F_L^{(s)},D_{pred}^{(s)}).
\end{aligned}
\end{equation}

Here, $f_{unc}^{(s)}$ denotes the learnable uncertainty
predictor, while $f_{off}^{(s)}$ denotes the residual
offset predictor. The offset predictor $f_{off}^{(s)}$
employs two $3\times3$ convolutions followed by
two $1\times1$ convolutions.
For training $U^{(s)}$, since ground-truth uncertainty
labels are unavailable, pseudo-targets are generated
from the propagated disparity error:

\begin{equation}
U_{gt}^{(s)}
=
1-\exp\left(
-\beta
\frac{
\max\left(0,
|D_{pred}^{(s)}-D_{gt}^{(s)}|-R^{(s)}
\right)
}{
R^{(s)}
}
\right),
\label{eq:uncertainty_target}
\end{equation}

where $D_{gt}^{(s)}$ denotes ground-truth disparity,
$R^{(s)}$ is the local search radius, and $\beta$
controls uncertainty sensitivity. The target is zero
when the disparity error is within the search radius
and approaches one as the error exceeds it.

The predicted uncertainty modulates the initial
offset to adjust the local search center:

\begin{equation}
\begin{aligned}
\Delta O_{final}^{(s)}
&=U^{(s)}\cdot\Delta O_{init}^{(s)},\\
D_{center}^{(s)}
&=D_{pred}^{(s)}+\Delta O_{final}^{(s)}.
\end{aligned}
\end{equation}

Local disparity candidates are then sampled around
the adjusted center:

\begin{equation}
\mathcal{D}^{(s)}
=
\left\{
D_{center}^{(s)}+k\delta^{(s)}
\;\middle|\;
k\in[-r,r]\cap\mathbb{Z}
\right\},
\end{equation}

where $r$ is the sampling radius and $\delta^{(s)}$
denotes the disparity sampling interval.

UGRSM operates at the $1/8$ and $1/4$ resolution
stages, while the $1/16$ stage uses a full cost
volume. This enables uncertainty-guided search
center correction without expanding the local
search range.

\subsection{Mono-Stereo Mutual Refinement}

At the final $\frac{1}{4}$ resolution stage,
TRUST-Stereo follows the mono-stereo mutual
refinement strategy of MonSter~\cite{cheng2025monster}.
Stereo and monocular disparity estimates are used
to warp right-image features toward the left view,
generating complementary feature-difference cues.
These cues, together with geometric features and disparity estimates, enable iterative stereo
refinement and reciprocal monocular refinement.

\subsection{Training Objective}

TRUST-Stereo is trained using disparity, uncertainty,
and feature-level distillation supervision.
The stereo and uncertainty losses are defined as

\begin{equation}
\begin{aligned}
\mathcal{L}_{stereo}
&= \sum_{i=1}^{N}
\gamma^{N-i}\|D_i-D_{gt}\|_1,\\
\mathcal{L}_{unc}
&= \sum_{i=1}^{N_u}
\gamma^{N_u-i}\|U_i-U_{gt}\|_1,
\end{aligned}
\end{equation}

where $\gamma$ assigns greater weight to later
predictions, and $U_{gt}$ is defined in
Eq.~\eqref{eq:uncertainty_target}.

During the second training stage, feature-level
distillation is additionally applied. The overall
training objective is

\begin{equation}
\mathcal{L} =
\mathcal{L}_{stereo}
+\lambda_{unc}\mathcal{L}_{unc}
+\lambda_{KD}\mathcal{L}_{KD},
\end{equation}

where $\lambda_{unc}$ and $\lambda_{KD}$ balance
the uncertainty and distillation losses.

% \begin{figure}[!t]
%     \centering
%     \includegraphics[
%         width=\columnwidth,
%         height=0.36\textheight,
%         keepaspectratio
%     ]{figures/ETH3D_compare.pdf}
%     \caption{Qualitative zero-shot comparison on ETH3D.
%     Disparity predictions from BANet-2D, LightStereo-M,
%     and TRUST-Stereo (ours) are shown alongside the input
%     images. All models are trained exclusively on SceneFlow
%     without target-domain fine-tuning.}
%     \label{fig:eth3d_comparison}
% \end{figure}

\section{Experiment}
\begin{table}[t]
\centering

\caption{Zero-shot generalization on KITTI 2012,
KITTI 2015, ETH3D, and Middlebury-H.
Baseline results are adopted from~\cite{jing2026lite},
except Fast Foundation Stereo$^{\mathrm{R}}$,
which we retrained on SceneFlow since
the official SceneFlow-only weights are not released.
ETH3D and Middlebury-H results are evaluated
on valid non-occluded pixels.
SceneFlow-only methods are trained on SceneFlow.
In million-scale training, $^*$ denotes models retrained
using the synthetic data and training strategy
of~\cite{guo2024stereoanything};
$\dagger$ denotes 30M pseudo-labeled samples;
$\ddagger$ denotes 1.8M synthetic and 0.5M
real-world pseudo-labeled samples;
and $\S$ denotes our 1.4M synthetic-only training.
Best and second-best results within each setting
are highlighted in red (bold) and yellow, respectively.}

\label{tab:benchmark}

% Reduce column spacing so the table can be rendered with a larger font
\setlength{\tabcolsep}{2.5pt}

\resizebox{\linewidth}{!}{
\begin{tabular}{lccccccc}
\toprule

% \multirow{Method} 
% & \multicolumn{2}{c}{KITTI 2012} 
% & \multicolumn{2}{c}{KITTI 2015}
% & \multicolumn{1}{c}{ETH3D}
% & \multicolumn{1}{c}{Middlebury}
% & \multirow{Runtime (ms)} \\

\multirow{2}{*}{Method}
& \multicolumn{2}{c}{KITTI 2012}
& \multicolumn{2}{c}{KITTI 2015}
& \multicolumn{1}{c}{ETH3D}
& \multicolumn{1}{c}{Middlebury}
& \multirow{2}{*}{Runtime (ms)} \\

\cmidrule(lr){2-3}
\cmidrule(lr){4-5}
\cmidrule(lr){6-6}
\cmidrule(lr){7-7}

& D1 
& EPE 
& D1 
& EPE
& Bad 1.0
& Bad 2.0 
& \\

\midrule

\multicolumn{8}{l}{\textit{Efficient methods: SceneFlow training}} \\
\midrule

CoEX~\cite{bangunharcana2021correlate}
& 22.30 & 3.37
& 17.33 & 3.00
& 31.97
& 26.42
& 27 \\

MobileStereoNet-2D~\cite{shamsafar2022mobilestereonet}
& 19.30 & 2.51
& 21.88 & 2.85
& 17.10
& 37.98
& 41 \\

MobileStereoNet-3D~\cite{shamsafar2022mobilestereonet}
& 19.59 & 2.67
& 17.89 & 3.01
& 18.42
& 25.26
& 67 \\

FastACV+~\cite{xu2024accurate}
& 16.69 & 2.29
& 15.28 & 3.27
& 11.50
& 27.34
& 45 \\

LightStereo-M~\cite{guo2025lightstereo}
& 6.76 & 1.26
& 6.79 & 1.42
& 13.93
& 16.99
& \cellcolor{yellow!40}20 \\

LightStereo-L~\cite{guo2025lightstereo}
& 6.80 & 1.27
& 6.62
& \cellcolor{yellow!40}1.29
& 9.66
& 17.23
& 34 \\

BANet-2D~\cite{xu2025banet}
& 14.75 & 2.23
& 16.98 & 3.91
& 44.89
& 26.79
& 21 \\

BANet-3D~\cite{xu2025banet}
& 17.39 & 2.41
& 17.28 & 3.36
& 29.27
& 28.79
& 30 \\

Lite Any Stereo~\cite{jing2026lite}
& \cellcolor{yellow!40}5.45
& 1.18
& \cellcolor{yellow!40}6.45
& 1.32
& 15.38
& \cellcolor{yellow!40}{13.13}
& \cellcolor{red!25}\textbf{7} \\

WAFT-Stereo (DAv2S, 4)~\cite{wang2026waftstereo}
& 9.78
& --
& 16.60
& --
& 71.10
& --
& 47 \\

Fast Foundation Stereo$^{\mathrm{R}}$~\cite{wen2026fastfoundationstereo}
&
6.29
&
\cellcolor{yellow!40}1.06
&
7.06
&
1.50
&
 \cellcolor{red!25}\textbf{6.36}
&
16.51
& 
41
\\

\midrule

\textbf{TRUST-Stereo (Ours)}
& \cellcolor{red!25}\textbf{4.30}
& \cellcolor{red!25}\textbf{0.93}
& \cellcolor{red!25}\textbf{5.78}
& \cellcolor{red!25}\textbf{1.16}
& \cellcolor{yellow!40} 8.79
& \cellcolor{red!25}\textbf{10.14}
& 50 \\

\midrule

\multicolumn{8}{l}{\textit{Efficient methods: Million-scale training}} \\
\midrule

LightStereo-M$^*$~\cite{guo2025lightstereo}
&4.10  & 0.99
& 4.97 & 1.13
& 5.33
& 10.58
& \cellcolor{yellow!40}{20}\\

BANet-2D$^*$~\cite{xu2025banet}
& 3.90 & 0.93
&4.71  & 1.07
& 5.92
& 10.05
& 21 \\

StereoAnything-L$^\dagger$~\cite{guo2024stereoanything}
& 4.00& 0.92
&4.81  & 1.10
& 3.81
& 9.82
& -- \\

Lite Any Stereo$^\ddagger$~\cite{jing2026lite} 
& \cellcolor{red!25}\textbf{3.04} & 0.79
& 3.87 & 0.99
& 3.53
& \cellcolor{yellow!40}7.51
& \cellcolor{red!25}\textbf{7}\\

\textbf{TRUST-Stereo$^\S$ (Ours)}
& \cellcolor{yellow!40}{3.20}
& \cellcolor{red!25}\textbf{0.77}
& \cellcolor{red!25}\textbf{3.49}
& \cellcolor{red!25}\textbf{0.91}
& \cellcolor{red!25}\textbf{2.13}
&  \cellcolor{red!25}\textbf{6.79}
&  50\\

\bottomrule
\end{tabular}
}

\end{table}

\begin{table*}[!t]
\centering
\caption{Fine-tuned comparison on the KITTI 2012 and
KITTI 2015 test sets. Lower values indicate better
performance. Baseline MACs are adopted
from~\cite{jing2026lite} and reported at
$1242\times375$ input resolution.
}
\label{tab:kitti_finetuned}

\small
\setlength{\tabcolsep}{3pt}
\renewcommand{\arraystretch}{1.05}

\begin{tabular}{lccccc|ccc|c}
\toprule
Method
& \multicolumn{5}{c|}{KITTI 2012}
& \multicolumn{3}{c|}{KITTI 2015}
& MACs (G) \\

\cmidrule(lr){2-6}
\cmidrule(lr){7-9}

& 3-noc
& 3-all
& 4-noc
& 4-all
& EPE-noc / all
& D1-bg
& D1-fg
& D1-all
& \\

\midrule

DispNetC~\cite{mayer2016dispnet}
& 4.11 & 4.65 & 2.77 & 3.20 & 0.9 / 1.0
& 4.32 & 4.41 & 4.34 & -- \\

DeepPruner-Fast~\cite{duggal2019deeppruner}
& -- & -- & -- & -- & --
& 2.32 & 3.91 & 2.59 & 194 \\

AANet+~\cite{xu2020aanet}
& 1.55 & 2.04 & 1.20 & 1.58 & 0.4 / 0.5
& 1.65 & 3.96 & 2.03 & -- \\

DecNet~\cite{yao2021decnet}
& -- & -- & -- & -- & --
& 2.07 & 3.87 & 2.37 & -- \\

BGNet+~\cite{xu2021bgnet}
& 1.62 & 2.03 & 1.16 & 1.48 & 0.5 / 0.6
& 1.81 & 4.09 & 2.19 & 77 \\

HITNet~\cite{tankovich2021hitnet}
& 1.41 & 1.89 & 1.14 & 1.53 & 0.4 / 0.5
& 1.74 & 3.20 & 1.98 & 47 \\

CoEx~\cite{bangunharcana2021correlate}
& 1.55 & 1.93 & 1.15 & 1.42 & 0.5 / 0.5
& 1.79 & 3.82 & 2.13 & 54 \\

MobileStereoNet-2D~\cite{shamsafar2022mobilestereonet}
& -- & -- & -- & -- & --
& 2.49 & 4.53 & 2.83 & 127 \\

MobileStereoNet-3D~\cite{shamsafar2022mobilestereonet}
& -- & -- & -- & -- & --
& 2.75 & 3.87 & 2.10 & 564 \\

Lite-CREStereo++~\cite{jing2023uncertainty}
& 1.43 & 1.82 & 1.12 & 1.44 & 0.5 / 0.5
& 1.79 & 3.53 & 2.08 & 93 \\

LightStereo-M~\cite{guo2025lightstereo}
& 1.56 & 1.91 & 1.10 & 1.36 & 0.5 / 0.5
& 1.81 & 3.22 & 2.04 & 33 \\

LightStereo-L~\cite{guo2025lightstereo}
& 1.55 & 1.87 & 1.10 & 1.33 & 0.5 / 0.5
& 1.78 & 2.64 & 1.93 & 84 \\

BANet-2D~\cite{xu2025banet}
& 1.38 & 1.79 & 1.01 & 1.32 & 0.5 / 0.5
& 1.59 & 3.03 & 1.83 & 36 \\

BANet-3D~\cite{xu2025banet}
& 1.27 & 1.72 & 0.95 & 1.27 & 0.5 / 0.5
& 1.52 & 3.02 & 1.77 & 78 \\

Lite Any Stereo~\cite{jing2026lite} &
\textbf{1.09} & \textbf{1.49} & \textbf{0.76} & \textbf{1.04} & \textbf{0.4 / 0.5} &
1.36 & 3.45 & \textbf{1.71} & 33 \\
\midrule

\textbf{TRUST-Stereo (Ours)}
& 1.15 & 1.52
& 0.87 & 1.12
& \textbf{0.4 / 0.5}
&\textbf{1.42} &\textbf{2.74} & 1.64 & 291 \\

\bottomrule
\end{tabular}

\end{table*}

\subsection{Zero-Shot Generalization}

\textbf{Implementation Details.}
TRUST-Stereo is trained under two settings: SceneFlow-only
and million-scale training with approximately 1.4M synthetic
stereo pairs from SceneFlow~\cite{mayer2016dispnet},
TartanAir~\cite{wang2020tartanair},
FallingThings~\cite{tremblay2018fallingthings},
Virtual KITTI 2~\cite{cabon2020vkitti2},
FoundationStereo~\cite{wen2025foundationstereo},
and Dynamic Replica~\cite{karaev2023dynamicstereo}. We use maximum disparity of 192
during training. All zero-shot evaluations are reported
without target-domain fine-tuning.

\textbf{Benchmark Datasets and Metrics.}
We evaluate on KITTI 2012, KITTI 2015, ETH3D, and
Middlebury-H using EPE, D1, Bad 1.0, and Bad 2.0.
All results are reported on valid non-occluded pixels.

\begin{figure*}[!t]
    \centering
    \includegraphics[
        width=0.95\textwidth,
        keepaspectratio
    ]{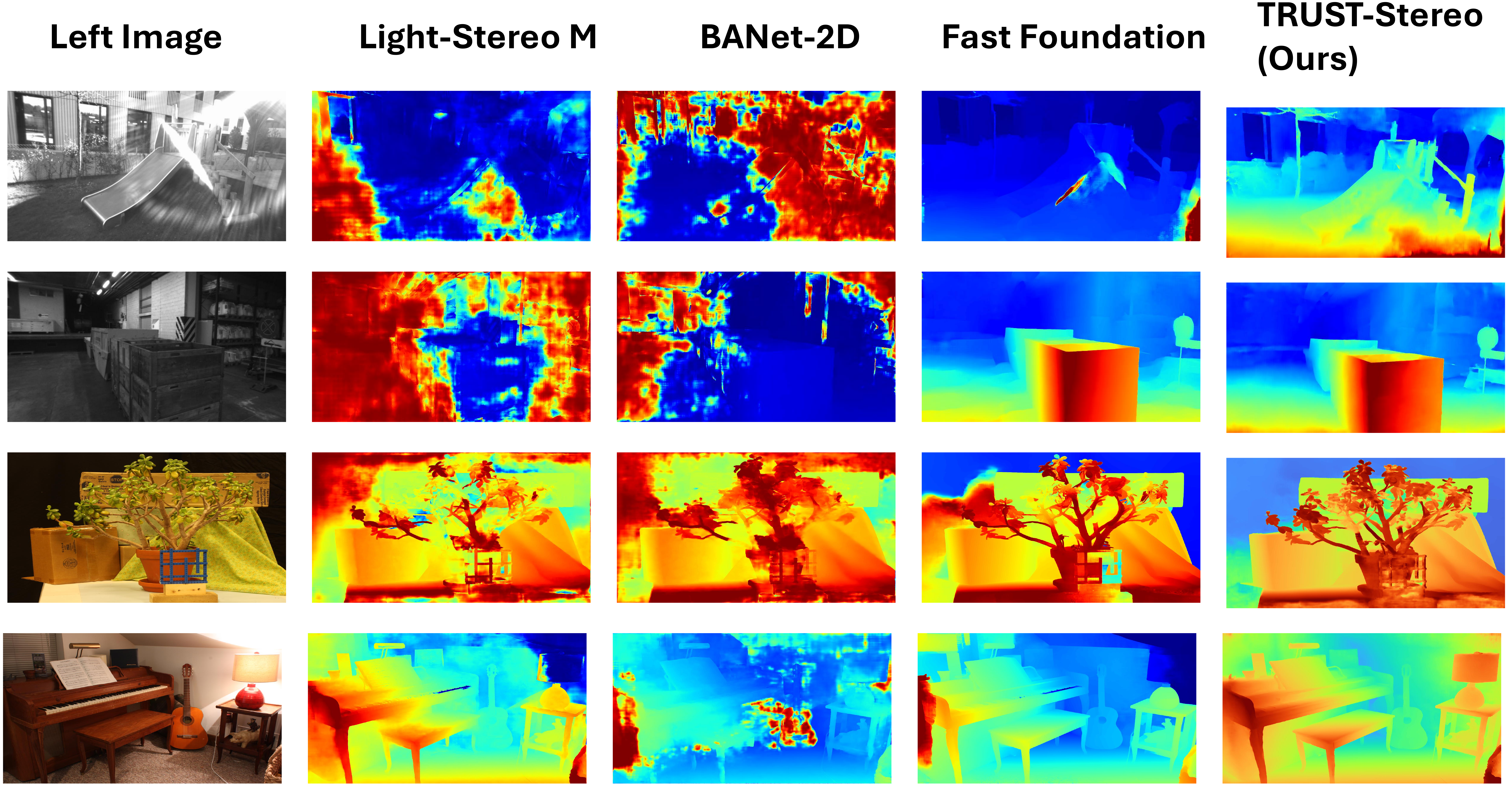}

    \caption{Qualitative zero-shot disparity estimation
    comparison on ETH3D and Middlebury-H.
    The first two rows show ETH3D, while the last two rows
    show Middlebury-H. Disparity predictions from
    TRUST-Stereo (ours), Fast-FoundationStereo,
    LightStereo-M, and BANet-2D are compared.
    All models are trained exclusively on SceneFlow
    and evaluated zero-shot without target-domain fine-tuning.}
    
    \label{fig:comparision}
\end{figure*}

\textbf{Quantitative Comparison.}
As shown in Tab.~\ref{tab:benchmark}, compared to the state-of-the-art efficent stereo matching methods, TRUST-Stereo
achieves the best zero-shot training results on five of six metrics under
SceneFlow-only training. Qualitative comparisons on
ETH3D and Middlebury against Fast Foundation Stereo,
BANet-2D, and LightStereo-M are presented in
Fig.~\ref{fig:comparision}.
Under million-scale training, using only 1.4M synthetic
stereo pairs without real-world pseudo-labeled data,
TRUST-Stereo achieves the best performance on all six
metrics, including KITTI 2012 EPE (0.77), KITTI 2015
D1 (3.49\%) and EPE (0.91), ETH3D Bad 1.0 (2.13\%),
and Middlebury Bad 2.0 (6.79\%). These results demonstrate
strong zero-shot generalization with less training data.

\subsection{Fine-Tuned Benchmark Results}

% Although benchmark-specific fine-tuning is not the primary focus of this work, we additionally evaluate TRUST-Stereo in this setting for completeness. The pretrained model is fine-tuned on the KITTI 2012 and KITTI 2015 training sets and evaluated on the corresponding official test sets. We compare against representative efficient stereo matching methods using standard benchmark metrics, with results reported in Tab.~\ref{tab:kitti_finetuned}.

Although benchmark-specific fine-tuning is not the primary
focus of this work, we additionally evaluate TRUST-Stereo
in this setting for completeness. The pretrained model is
fine-tuned on the KITTI 2012 and KITTI 2015 training sets
and evaluated on the corresponding official test sets.
We compare against representative efficient stereo matching
methods using standard benchmark metrics, with results
reported in Tab.~\ref{tab:kitti_finetuned}.
On KITTI 2012, TRUST-Stereo achieves competitive results,
with 3-noc and 3-all errors of 1.15\% and 1.52\%,
respectively, approaching the performance of Lite Any
Stereo, which achieves 1.09\% and 1.49\%.
On KITTI 2015, TRUST-Stereo achieves the best D1-fg
and D1-all results among the compared methods,
reducing D1-fg from 3.45\% to 2.74\% and D1-all
from 1.71\% to 1.64\% compared with Lite Any Stereo.
Notably, TRUST-Stereo is pretrained using only
1.4M synthetic stereo pairs, whereas Lite Any Stereo
benefits from a larger and more diverse pretraining
dataset that includes real-world samples.
These results demonstrate the effectiveness of TRUST-Stereo
under benchmark-specific fine-tuning, complementing
its strong zero-shot generalization performance.

\begin{table}[t]
\centering
\caption{Effect of UGRSM on local cost-volume coverage under
SceneFlow-only zero-shot generalization. Values indicate the
percentage of ground-truth correspondences outside the local
search range. Lower is better.}
\label{tab:ugrsm_coverage}

\footnotesize
\setlength{\tabcolsep}{2.5pt}

\begin{tabular}{lcc}
\toprule

Dataset & w/o UGRSM (\%) & w/ UGRSM (\%) \\

\midrule

KITTI 2012
& 0.2833
& \textbf{0.2701} \\

KITTI 2015
& 0.5189
&\textbf{0.3188} \\

ETH3D
& 2.7000
& \textbf{1.4300} \\

Middlebury
& 4.0777
& \textbf{2.6741} \\

\bottomrule
\end{tabular}

\end{table}

\subsection{Ablation Study}

We conduct ablation studies to separately evaluate the
effectiveness of the proposed UGRSM and SJTA
knowledge distillation strategy. Both components are evaluated
under SceneFlow-only training, with zero-shot generalization
assessed on KITTI 2012, KITTI 2015, ETH3D, and Middlebury.

\noindent\textbf{Effect of Uncertainty-Guided Residual Search.}
We evaluate UGRSM by comparing models trained with and without the proposed module. Knowledge distillation is disabled in both settings to isolate the contribution of UGRSM. Both models are trained exclusively on SceneFlow and evaluated zero-shot on four unseen datasets.

As shown in Table~\ref{tab:ugrsm_coverage} , UGRSM consistently reduces the percentage of valid ground-truth correspondences falling outside the local cost-volume search range. Specifically, the out-of-range rate decreases from 0.2833\% to 0.2701\% on KITTI 2012, 0.5189\% to 0.3188\% on KITTI 2015, 2.7000\% to 1.4300\% on ETH3D, and 4.0777\% to 2.6741\% on Middlebury. These correspond to relative reductions of 4.7\%, 38.6\%, 47.0\%, and 34.4\%, respectively.

These results demonstrate that UGRSM improves local search coverage by adaptively correcting inaccurate disparity propagation, with particularly substantial improvements on ETH3D and Middlebury under zero-shot generalization.

\begin{table}[!t]
\centering

\caption{Effect of uncertainty-guided residual search on
zero-shot disparity accuracy. All variants are trained
on SceneFlow without knowledge distillation.
ETH3D and Middlebury are evaluated on all valid pixels,
including occluded regions. Lower is better.}

\label{tab:ugrsm_accuracy}

\setlength{\tabcolsep}{3pt}
\small

\resizebox{\columnwidth}{!}{
\begin{tabular}{llccc}
\toprule
Dataset & Metric
& \shortstack{w/o\\UGRSM}
& \shortstack{UGRSM\\(No Uncertainty)}
& \shortstack{Full UGRSM\\(Ours)} \\
\midrule

KITTI 2012
& D1 (\%)
& 4.81
& 4.85
& \textbf{4.64} \\

\midrule

KITTI 2015
& D1 (\%)
& 6.79
& 6.73
& \textbf{6.23} \\

\midrule

ETH3D
& Bad 1.0 (\%)
& 9.62
& 13.46
& \textbf{9.12} \\

\midrule

Middlebury
& Bad 2.0 (\%)
& 15.96
& 14.82
& \textbf{14.62} \\

\bottomrule
\end{tabular}
}
\end{table}
As shown in Tab.~\ref{tab:ugrsm_accuracy}, UGRSM
reduces KITTI 2015 D1 from 6.79\% to 6.23\%,
while improving KITTI 2012 D1 from 4.81\% to 4.64\%.
ETH3D Bad 1.0 decreases from 9.62\% to 9.12\%,
and Middlebury Bad 2.0 decreases from 15.96\% to 14.62\%.

\textbf{Effect of Knowledge Distillation.}
We compare the proposed SJTA
strategy against direct Teacher--Student KD and ViTKD,
with UGRSM enabled in all variants. All models are trained
exclusively on SceneFlow and evaluated zero-shot on KITTI 2012,
KITTI 2015, ETH3D, and Middlebury. As shown in
Tab.~\ref{tab:ablation_kd_generalization}, this comparison
assesses the effectiveness of intermediate teacher-assistant
representations relative to direct and ViT-specific feature
distillation.

SJTA achieves the best results on nine of ten metrics.
The largest improvements occur on ETH3D, where Bad 1.0
decreases from 24.80\% (direct KD) and 21.31\% (ViTKD)
to 8.79\%, while Bad 2.0 decreases from 20.43\% and
16.55\% to 4.35\%, respectively. Improvements are also
observed on Middlebury and KITTI, except for KITTI 2015
D1, where direct KD performs slightly better.
These results demonstrate the effectiveness of SJTA
for cross-domain feature transfer.

\begin{table}[!t]
\centering
\caption{Zero-shot comparison of direct Teacher--Student KD,
ViTKD, and the proposed Senior--Junior TA KD.
Lower is better.}
\label{tab:ablation_kd_generalization}
\resizebox{\linewidth}{!}{
\begin{tabular}{llccc}
\toprule
Dataset & Metric & Direct KD & ViTKD & SJTA KD (Ours) \\
\midrule
KITTI 2012 & EPE & 0.95 & 0.99 & \textbf{0.93} \\
           & D1 (\%) & 4.45 & 4.96 & \textbf{4.30} \\
\midrule
KITTI 2015 & EPE & 1.19 & 1.18 & \textbf{1.16} \\
           & D1 (\%) & \textbf{5.71} & 6.35 & 5.78 \\
\midrule
ETH3D & Bad 1.0 (\%) & 24.80 & 21.31 & \textbf{8.79} \\
      & Bad 2.0 (\%) & 20.43 & 16.55 & \textbf{4.35} \\
      & Bad 3.0 (\%) & 18.68 & 14.51 & \textbf{2.74} \\
\midrule
Middlebury & Bad 1.0 (\%) & 25.92 & 21.86 & \textbf{19.07} \\
           & Bad 2.0 (\%) & 15.82 & 13.48 & \textbf{10.14} \\
           & Bad 3.0 (\%) & 11.50 & 10.34 & \textbf{7.06} \\
\bottomrule
\end{tabular}
}
\end{table}

\textbf{Joint Contribution of SJTA and UGRSM.} To investigate the individual and combined effects of the proposed components, we conduct a four-way ablation under SceneFlow-only training. As shown in Tab. \ref{tab:ablation_sjta_ugrsm}, both SJTA and UGRSM independently improve zero-shot generalization over the baseline. Combining both components achieves the best performance, reducing D1 from 4.81\% to 4.30\% on KITTI 2012 and from 6.79\% to 5.78\% on KITTI 2015. These results demonstrate the complementary benefits of teacher-assisted feature distillation and uncertainty-guided disparity search.

\begin{table}[t]
\centering
\caption{Ablation study of the proposed Senior--Junior Teacher
Assistant (SJTA) distillation and Uncertainty-Guided Residual
Search Module (UGRSM) under SceneFlow-only training.
All variants are evaluated zero-shot on KITTI 2012 and
KITTI 2015 using D1 (\%). Lower is better.}
\label{tab:ablation_sjta_ugrsm}
\begin{tabular}{cc|cc}
\toprule
SJTA & UGRSM & KITTI 2012 & KITTI 2015 \\
 & & D1 (\%) $\downarrow$ & D1 (\%) $\downarrow$ \\
\midrule
$\times$ & $\times$ & 4.81 & 6.79 \\
$\times$ & $\checkmark$ & 4.64 & 6.23 \\
$\checkmark$ & $\times$ & 4.60 & 6.43 \\
$\checkmark$ & $\checkmark$ & \textbf{4.30} & \textbf{5.78} \\
\bottomrule
\end{tabular}
\end{table}

\FloatBarrier

\section{Conclusion}

We presented TRUST-Stereo, a real-time coarse-to-fine stereo
matching framework that improves disparity refinement and
zero-shot generalization. The proposed Uncertainty-Guided
Residual Search Module (UGRSM) estimates the reliability
of propagated disparities and adaptively adjusts local
cost-volume centers to recover out-of-range correspondences.
Additionally, our two-stage Senior--Junior Teacher Assistant
(SJTA) strategy transfers rich representations from a large
ViT-L teacher to a lightweight ViT-S backbone without
requiring costly large-scale assistant pretraining.

Experiments on KITTI 2012, KITTI 2015, ETH3D, and Middlebury
demonstrate strong zero-shot generalization using only 1.4M
synthetic stereo pairs, outperforming competing efficient
methods on several metrics while maintaining real-time
inference at approximately 50 ms per stereo pair.
Ablation studies further confirm the contributions of
UGRSM and SJTA to robust stereo matching.

\section*{LLM Usage Statement}
OpenAI's ChatGPT was used to assist with manuscript
editing, LaTeX formatting, and code development and
debugging. All scientific contributions, experimental
results, and conclusions were reviewed and verified
by the authors.

\bibliographystyle{ieeenat_fullname}
\bibliography{main}

{
    \small
    \bibliographystyle{ieeenat_fullname}
    \bibliography{main}
}


\end{document}